\documentclass[runningheads]{llncs}
\usepackage{graphicx}
\usepackage{gensymb}
\usepackage{array}
\usepackage{makecell}
\usepackage{amsmath}
\usepackage{subfig}
\usepackage{hyperref}
\usepackage{todonotes}
\usepackage{float}

\begin{document}
\title{Rhoban Football Club: RoboCup Humanoid Kid-Size 2023 Champion Team Paper}

\titlerunning{Rhoban Football Club: RoboCup Humanoid Kid-Size 2023 Champion Team Paper}
% If the paper title is too long for the running head, you can set
% an abbreviated paper title here
%
\author{
  Julien Allali \and
  Adrien Boussicault \and
  Cyprien Brocaire \and
  C\'eline Dobigeon \and
  Marc Duclusaud \and
  Cl\'ement Gaspard \and
  Hugo Gimbert \and
  Lo\"{i}c Gondry \and
  Olivier Ly \and
  Gr\'egoire Passault \and
  Antoine Pirrone
}
\authorrunning{Rhoban Football Club}
\institute{
    Rhoban Football Club Team,\\
    LaBRI, University of Bordeaux, France\\
    \email{team@rhoban.com}\\
    Corresponding author: Grégoire Passault\\
    \email{gregoire.passault@u-bordeaux.fr}
}
\maketitle              % typeset the header of the contribution
\begin{abstract}
  In 2023, Rhoban Football Club reached the first place of the KidSize soccer
  competition for the fifth time, and received the \textit{best humanoid award}.
  This paper presents and reviews important points in robots architecture and workflow,
  with hindsights from the competition.
\end{abstract}
\section*{Introduction}

During the 2023 RoboCup kid-size soccer competition, Rhoban played 8 games, scored 95 goals,
and took a total of 2 goals (both during the final against CIT Brains, Japan), reached the first
place for drop-in games, technical challenges and software challenge. Those achievements resulted
in obtaining the \textit{best humanoid award} for Sigmaban, Rhoban's kid-size humanoid robot. \\
This paper presents the main aspects of our robots, focusing on the recent improvements.
Section \ref{sec:meca} gives a short overview of the robots architecture, along with recent mechanical
upgrades, section \ref{sec:vision} presents our vision and perception system, section \ref{sec:walk}
details our new walk engine and section \ref{sec:viewer_simulator} describes the strategy and
monitoring system.

\section{Robots architecture}
\label{sec:meca}

\subsection{Overview}

Sigmaban has 20 degrees of freedom, 3 per arm (MX-64), 2 for the head (MX-64) and 6 per leg (1xMX-64 + 5xMX-106).
The legs are fully serial (no parallel kinematics). Parts are mostly milled in 2017 aluminium alloy.
They are equipped with Intel NUC (i5) running a (non real-time) Debian and a low-level custom board with STM32 (72Mhz)
to interface devices communication with the computer.
We use BNO055 IMU (Accelerometer+Gyroscope) and each foot is equipped with four
strain jauges to measure the pressure. More details can be found in the public team description paper and
specifications
\footnote{
    \url{https://humanoid.robocup.org/previous-editions/robocup-2023/teams/}
}. \\
OnShape\footnote{\url{https://www.onshape.com/}} is used as CAD software, which allows us a seamless
CAD-to-URDF conversion using OnShape-to-robot\cite{onshape_to_robot},
our own open-source tool (winner of the 2023 RoboCup software challenge).

\subsection{Software}

Most of the software running on the robots is written in C++. We factored out core components
such as motion planning, control and strategy to separate modules and using Python
bindings, allowing us to test, visualize,
plot, run physics simulation in a faster workflow than C++, while keeping the same code
running on the robot. We switched from RBDL\cite{felis2017rbdl} to \textit{pinocchio}\cite{carpentier2019pinocchio} as rigid body dynamics library,
because of its state-of-the-art performances and active development.

\subsection{Recent mechanical upgrades}

\subsubsection{Bumpers}

Shocks with other robots and the ground are very common during the games, and mechanical breaks are frequent
during the competition. Our previous design was relying on bent piano wires to dampen the shocks, which has
proven to be efficient, but was not easy to reproduce, cumbersome and could create some obstruction with
the vision. We switched to 3D (filament) printed TPU bumpers on the torso and the shoulders for 2023. This approach
led to similar damping during lab tests.

\subsubsection{Hip yaw bearings}

In the leg architecture, the hip yaw degrees of freedom are typically highly stressed radially. In our previous
design, the whole leg was fixed at this point with one unique (M3) screw, with no easy possibility to use
a passive horn. We added cross roller bearings to this joint, and significantly reduced the backlash
and possible screw loosening or damage.
\section{Vision, perception and localisation}
\label{sec:vision}

\subsection{Lenses and Field of view}

When selecting a lens for a robot camera, a natural trade-off arises between the field of view and the
resolution/distortion of the image.
Our previous lens
had a field of view of 68°(H), 54°(V), which required too many head movements to scan the field. 
We switched to a higher field of view, keeping the same image processed resolution (644x482) because we wanted 
to keep the same computational power in order to still process the images at 30fps in real-time.
Our new lens (BF5M2023S23C\footnote{\href{https://www.lensation.de/product/BF5M2023S23C/}{Technical Documentation of BF5M2023S23C Lens}})
has a field of view of 100°(H), 83°(V). The time taken for a look-around scan was reduced from 3.2s to
2s to cover the same observation space.
In the future, we can consider removing the \textit{head pan}, which is arguably useless in the current
architecture.

\subsection{Camera calibration}

Intrinsic calibration was done for each camera using ChArUco boards, enabling tangential
and radial distortions. We found better results by imposing $k_3 = 0$ for radial coefficients during the calibration.
Since we are using mobile robots, the extrinsic calibration is not relevant and is replaced with pose estimation
produced by whole body forward kinematics. \\
In hindsight, the distortion on the border of the image is still high, which makes some part of the image
barely usable.
We also notice significant reprojection errors that we countered by adjusting the optical center of the camera,
especially to ensure proper positioning when the ball is close to the robot. Even if we had satisfying reprojection
results with such adjustment, we need to investigate whether it is caused to extrinsic errors (e.g. bent parts)
or intrinsic errors (e.g. errors in intrinsic calibration).

\subsection{Image processing}

We have switched from an approach based on Multi-class ROI and classification\cite{gondry2019rhoban}
to the state-of-the-art end-to-end YoloV8\cite{Jocher_YOLO_by_Ultralytics_2023} algorithm
for object detection, in order to detect different points of interest (POI, cf. \ref{subsec:dataset}) in the image.
Camera extrinsic estimation allows for the projection of the POI in the world frame and feeding
a particle filter \cite{gondry2019rhoban} for localisation. \\
Thanks to pressure sensors and the robot kinematics model, we also use extensively the odometry
to adjust the localisation.

\subsubsection{Training:}

In order to train the neural network, we need to have a dataset of images with the features labeled. The HTC Vive auto-labeling
described in \cite{gondry2019rhoban} was not enough accurate to train the neural network. We manually 
labeled a set of 400 images to train a first version of the neural network in order to pre-label the rest of the dataset. Correcting such pre-labeled image was considered two times faster than labeling an image from scratch.

We used Label-Studio\cite{Label_Studio} as labeling tool because it is open-source, comes with a user-friendly
web-based interface that can be shared through the network.

Yolo is trained with our dataset (see section \ref{subsec:dataset}) based on the pre-trained YoloV8n weights (3.2M parameters). It takes 48 minutes on an Intel Core™ i7-9700K, 32Go of RAM, a Nvidia RTX 2070 and a SSD to train the 
neural network for 407 epochs

\subsubsection{On-robot inference:}

Robots are running Intel Core™ i5-7260U with integrated GPU, 8Go of RAM and a SSD.
YoloV8n is translated to an OpenVINO\footnote{
 \url{https://www.openvino.ai/}   
} IR model, which is then loaded and run on the robot.
We apply our own post-processing (Non-Maximum Suppression) to the output, and also some custom filtering
such as removing unrealistic bounding boxes based on extrinsic information. \\
The robot runs the neural network online at approximately 48fps on integrated GPU instead 
of 5fps on CPU only (without OpenVINO).

\label{subsec:dataset}
\subsection{Dataset}

\subsubsection{Images selection and data augmentation:}
Our dataset is composed of 1794 images. We arbitrarily split it into the usual three parts to evaluate the performance of 
a machine learning model: Train set : 1462 images (80\%) | Evaluation set : 166 images (10\%) | Test set : 166 images (10\%).

\begin{figure}[h]
    \begin{minipage}[c]{.48\linewidth}
        \centering
        \includegraphics[height=4.5cm]{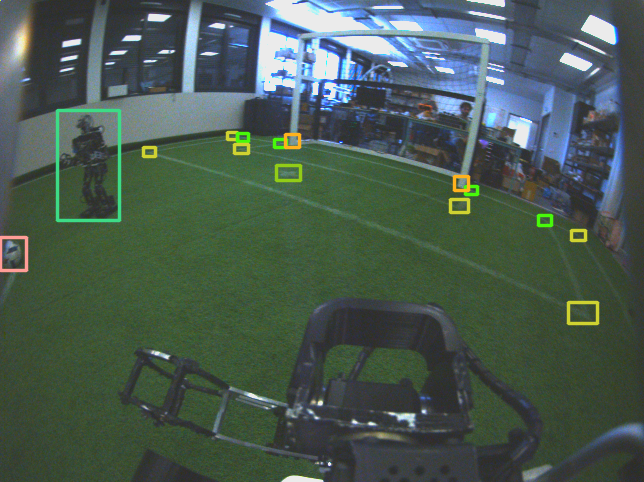}
        \caption{Inference on a frame with our trained neural network}
        \label{fig:example_yolo}
    \end{minipage}
    \hfill%
    \begin{minipage}[c]{.48\linewidth}
        \centering
        \includegraphics[trim={2cm 0 0 0},clip,height=5cm]{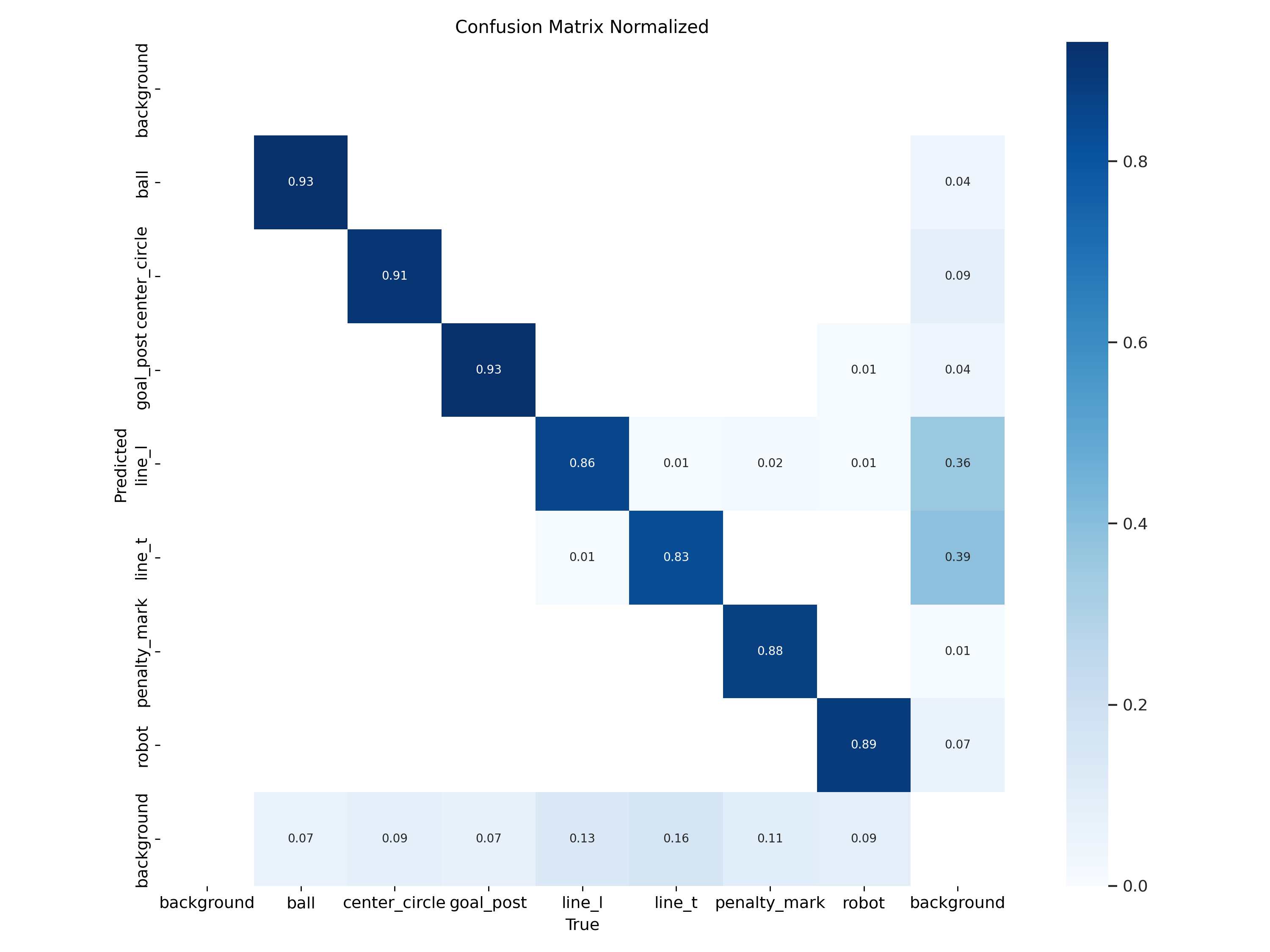}
        \caption{Normalized confusion matrix of our neural network (confidence threshold=0.25)}
        \label{fig:confusion_matrix_normalized}
    \end{minipage}
\end{figure}    

The majority of these images ($\sim$80\%) were captured at Rhoban lab with different brightness conditions and variations of the  shutter speed and white balance of the camera.
The rest of it was taken during the RoboCup 2022 and 2023 competition.
We had to train the network not only on images extracted from video (as in 2019\cite{gondry2019rhoban}) but also on raw images, 
because the network tends to learn compression artefacts and the onboard performances on the raw camera
stream was diminished. We also used data augmentation to make our network more robust with the following transformations: HSV modification, rotation, vertical flip, translation and rescaling.

\subsubsection{Detected features:}
We use 5 main localisation features:
The lines "L" and "T" patterns (4500 and 3500 instances), 
the center circle (700 its.), the penalty mark (800 its.) and the bottom of the goal posts (2000 its.). We also detect the ball (1500 its.) and the other robots 
(1500 its.).
Figure \ref{fig:example_yolo} displays an example of detection.

The normalized confusion matrix of the neural network training is depicted on figure \ref{fig:confusion_matrix_normalized}.
The L an T corners have a relatively high false positive rate (confused with background).
Note that this matrix is computed with a confidence threshold (0.25) lower than the one we use on the robot (0.6)
The rest of the matrix shows us that the neural network is able to detect the features with a high accuracy.

\section{Walk}
\label{sec:walk}

\subsection{Overview}

\begin{figure}[h]
\centering
\includegraphics[width=7cm]{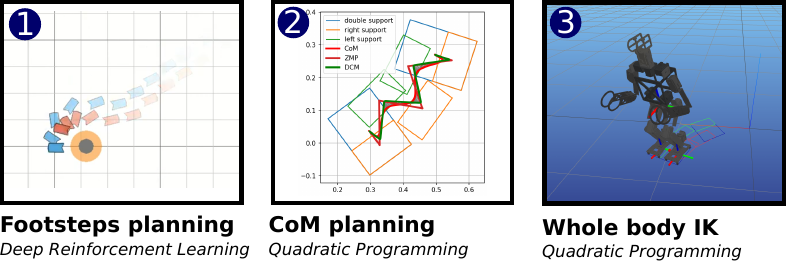}
\caption{Overview of the stages of the walk engine}
\label{fig:walk_overview}
\end{figure}

Our walk engine was almost entirely reworked in 2023.
The previous walk engine \cite{gondry2019rhoban} was based on cubic splines and
analytic inverse kinematics for the legs. This approach lacked a preview on the dynamics
of the center of mass and versatility in the tasks that can be used to define the
trajectories. This section describes the three main stages of this new walk engine,
as depicted on figure \ref{fig:walk_overview}: planning the footsteps,
planning the center of mass trajectory and finally computing the whole body inverse kinematics
to produce the joint trajectories that are sent to position-controlled actuators.
Note that kicking and standing-up are, however still achieved with hand-tuned keyframe animation, which
gives efficient results, but is tedious and not satisfying.

\label{sec:footsteps}

\subsection{Footsteps planning}

In the past, we relied on extensively hand-tuned approaches (based for example on potential fields) to plan
the footsteps.  On the one hand, this has proven to be efficient but relied on many hyper-parameters to be adjusted 
manually, especially when the robot walking abilities are updated, which is not satisfying.

Reinforcement learning, on the other hand, rely on simulated environments directly parametrized by the robot abilities along with a reward 
signal that can be adjusted to exactly match the desired outcome (in this case, minimizing the number of footsteps). 
It can embrace non-linearity and, thanks to Deep Reinforcement Learning and the use of neural networks as functions 
approximators, continuous actions and observations space to produce a policy with a fast on-line inference.
We thus formulated the footsteps planning as a reinforcement learning problem.

This RL environment is based on different parameters defined by our robot's structure: maximum step sizes 
for the robot's movement ($\Delta x_{max}$, $\Delta y_{max}$, and $\Delta \theta_{max}$), the feet spacing ($l_f$) 
and tolerances for reaching the target ($t_x$ and $t_\theta$ in meters and radians, respectively). 

The actions in this environment are represented as tuples of $(\Delta x, \Delta y, \Delta \theta)$, with the 
step sizes in meters and radians. These actions are clipped to ensure that they fall within an ellipsoid, 
rather than a box, to encourage more natural movement patterns. Footsteps are simulated using purely geometrical
displacements. the observation space consists of the position and orientation of the currently considered neutral 
support foot, along with information about the current support foot (left or right). These values are expressed in the 
target frame to represent "errors". 

The reward function in this RL environment is shaped as a cost function, which includes 
factors like the cost of taking additional steps, a reward shaping factor ($\alpha$), and penalties for collisions with the ball.

Technically, the environment is implemented as an OpenAI gym, and trained using
TD3\cite{fujimoto2018addressing} algorithm (implementation from \cite{raffin2019stable}). The policy is
then exported to OpenVINO IR model for inference, where the inference of one footstep takes about
$100 \mu s$ on the robot.

Extensive details about this work will be published later in a separate paper, and the environment is intended to
be open-sourced.

\subsection{CoM planning}

In order to ensure the robot stability, the center of mass is planned with a preview horizon,
on a scheme similar to \cite{kajita2003biped}.
This is formulated as the following discrete time optimization problem:

\begin{equation}
\label{eq:com_optimization}
\begin{split}
\min_{c, \dot c, \ddot c, \dddot c, z}
\space \space
&
\lVert z - z^d \rVert^2
+
\epsilon \lVert \dddot c \rVert^2
\\ 
s.t \space \space &
\begin{bmatrix}
c_{k+1} \\
\dot c_{k+1} \\
\ddot c_{k+1} \\
\end{bmatrix}
=
\begin{bmatrix}
1 & \Delta t & \frac{\Delta t^2}{2} \\
0 & 1 & \Delta t \\
0 & 0 & 1 \\
\end{bmatrix}
\begin{bmatrix}
c_{k} \\
\dot c_{k} \\
\ddot c_{k} \\
\end{bmatrix}
+
\begin{bmatrix}
\frac{\Delta t^3}{6} \\
\frac{\Delta t^2}{2} \\
\Delta t \\
\end{bmatrix}
\dddot c_{k}
\\
& z_k = c_k - \frac{h}{g} \ddot c_k, z_k \in S_k
\\
& c_0, \dot c_0, \ddot c_0 = c_{init}, \dot c_{init}, \ddot c_{init}
\\
& c_f, \dot c_f, \ddot c_f = c_{final}, \dot c_{final}, \ddot c_{final}
\end{split}
\end{equation}

Where:

\begin{itemize}
    \item
    $c_k, \dot c_k, \ddot c_k, \dddot c_k$ are respectively the (2D) center of mass position, velocity, acceleration and jerk at step $k$,
    \item
    $z_k$ is ZMP approximation under the LIPM model at step $k$, $h$ being the (assumed constant) height of the
    center of mass and $g$ the gravity,
    \item
    $z^d$ is a the desired ZMP trajectory,
    \item
    $S_k$ is the support polygon at step $k$, given by the current support footprint or by the convex hull of
    the two support footprints in the double support phase,
    \item
    Initial and final center of mass positions are given by the current state of the robot and the state at the
    end of the preview horizon.
\end{itemize}

The objective function is quadratic. The only constraint that is not an equality is $z \in S$,
which can be turned into inequalities because $S$ are convex polygons.
This can then be translated into a quadratic programming (QP)
problem. In practice, we used \textit{eiquadprog} \cite{eiquadprog} solver. Note that
$c, \dot c, \ddot c, \dddot c$ and $z$ are all coupled with equality constraints,
the problem can be rewritten (with less clarity) so that $\dddot c$ is the only decision variable.

The reference trajectory for the ZMP ($z^d$) is in our case a target that is constant in the current support
foot. This offset can then be tuned to adjust the walk, with for example the effect of increasing or reducing
the lateral swing. During the competition, this target was set to zero (i.e. we try to have the ZMP as
close as possible to the center of the support foot).

Because our single support duration was tuned to $360ms$, 
our value for $\Delta t$ is $36ms$, we plan $48$ steps ahead which implies $96$ decision variables
($\dddot c$) to optimize. The preview horizon is then $1.7s$ (roughly 5 walk steps).
Solving this optimization problem takes about $2ms$ on the robot and was done with a constant period of $25ms$.

Equation \ref{eq:com_optimization} can be integrated as a smooth $\mathcal{C}^2$ trajectory
for $c(t)$.

\subsection{Swing foot trajectory}

The swing foot trajectory is a trade-off between the risk of applying tangential forces on the ground
while taking off or landing and the need to limiting the speed that is required to reach the target.
Even if more thoughtful work can be considered in the future, we tried empirically different trajectories
and settled on the most efficient one.

\begin{figure}[h]
\centering
\includegraphics[width=8cm]{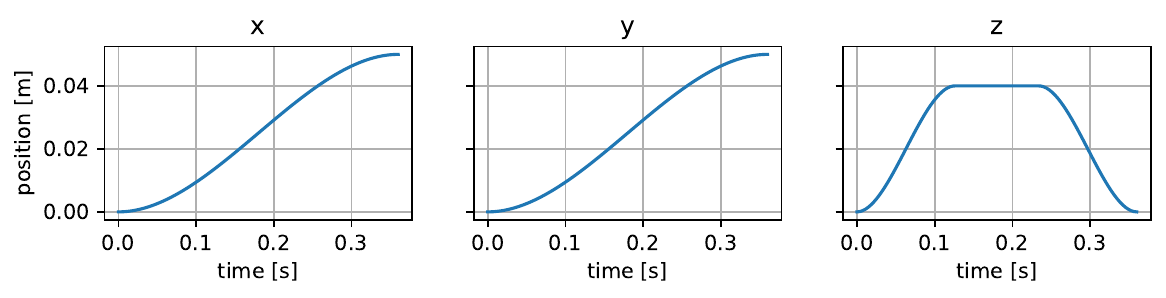}
\caption{Cartesian trajectory of a swinging foot of $5cm$ along $x$ axis and $5cm$ along $y$ axis}
\label{fig:swing_foot}
\end{figure}

$z$ being the vertical axis, $x(t)$ and $y(t)$ are simply cubic splines defined by initial and final
positions, and zero initial and final velocities (a similar approach is used for the yaw of the foot).
The vertical trajectory $z(t)$ is a piecewise
cubic splines with a taking off, a plateau (tuned to 30\% of the duration) and a landing phase
(see figure \ref{fig:swing_foot}).

\subsection{Whole body inverse kinematics}

The planning for the center of mass, along with the swing foot spline provides the following
task-space trajectories to be followed by the robot:

\begin{enumerate}
    \item
    \textbf{Center of mass}: output of the optimization problem from the previous stage and
    a constant height $h$,
    \item
    \textbf{Trunk orientation}: constant pitch/roll (tuned to leaning $11.5deg$ forward), and a yaw
    following the swing foot,
    \item
    \textbf{Support foot}: not moving in the world,
    \item
    \textbf{Swing foot}: following the spline described in the previous section and remaining
    parallel to the ground.
\end{enumerate}

Kinematically, task 1. is a position task (3D), task 2. an orientation task (3D) and tasks 3.
and 4. are frame tasks (6D). This results in assigning 18 degrees of freedom, which are exactly
the degrees of the legs (12) plus the floating base (6).

To find the corresponding joint-space trajectories $q(t)$, we need to solve the inverse kinematics
problem. Formally, tasks can be written as errors $e(q)$ that we want to take to zero.
Since $e(q)$ is highly non-linear, it is approximated by a first order Taylor expansion around the
current state $q_0$: $e(q) \approx e(q_0) + J(q_0) \Delta q$ where $J(q_0)$ is the Jacobian matrix
of $e$ at $q_0$. The problem is then to find $\Delta q$ such that $e(q_0) + J(q_0) \Delta q = 0$.
Again, we formulate this as a QP problem:

\begin{equation}
\begin{split}
\min_{\Delta q}
\space \space
&
\sum_i w_i \lVert J^s_i(q_0) \Delta q + e^s_i(q_0) \rVert^2
+
\epsilon \lVert \Delta q \lVert^2
\\ 
s.t \space \space &
J^h_i(q_0) \Delta_q + e^h_i(q_0) = 0
\\
& q_{min} < q_0 + \Delta q < q_{max}
\\
& - \Delta t . \dot q_{max} < \Delta q < \Delta t . \dot q_{max}
\end{split}
\end{equation}

Here:

\begin{itemize}
\item
$q_0$ is the current joint state of the (reference) robot, around which linear approximation
is made, and $\Delta q$ is the variation of the joint state that we want to find,
\item
$J^h$ and $e^h$ are the \textit{hard} tasks (equality constraints)
\item
$J^s$ and $e^s$ are the \textit{soft} tasks, ending up in the objective function to minimize
\item
$w_i$ are the weights of the soft tasks, that can be tuned to prioritize some tasks over others,
\item
$q_{min}$ and $q_{max}$ are the joint limits,
\item
$\dot q_{max}$ is the maximum joint velocity.
\end{itemize}

All the tasks are expressed in the world frame, and the Jacobian are computed locally for numerical
stability. The j$\epsilon \lVert \Delta q \lVert^2$ term is a regularization term that is used to
ensure that the solution is not too far from the current state, and ensure that the problem is
well-posed.

In practice, we also used \textit{eiquadprog} \cite{eiquadprog} solver, relying on
\textit{pinocchio} \cite{carpentier2019pinocchio} to compute efficiently the transformation
matrices and Jacobians from the URDF model that is kept consistent with CAD updates tanks
to Onshape-To-Robot\cite{onshape_to_robot}.
This optimization problem was solved at the control frequency (about $200Hz$, $\Delta t = 5ms$),
and takes approximately $100 \mu s$ on the robot.

\subsection{Open-source}

Efforts were made to get the previously mentioned optimization problems convenient to
formulate and solve from a software engineering point of view. Our mid-term objective is
to document and open-source all the parts of the walk engine.

\subsection{Limitations}

\begin{figure}[h]
\centering
\includegraphics[width=\textwidth]{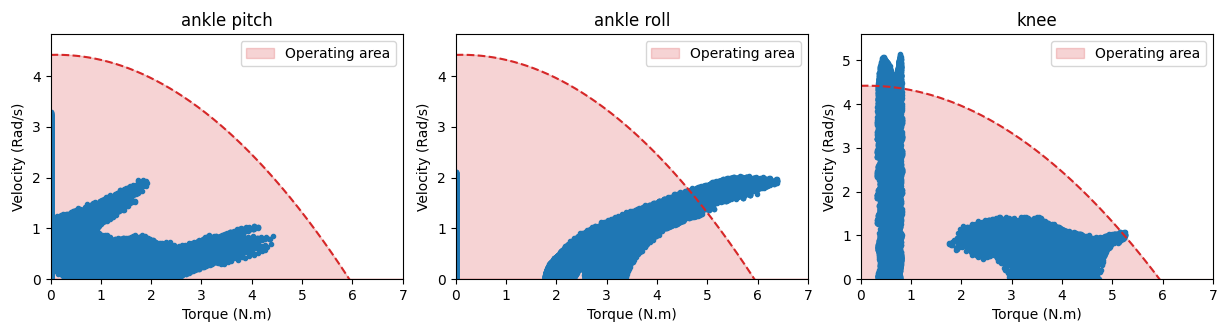}
\caption{Planned velocities during the walk for the knee and ankle actuators. The red area represent the operational range of the actuators given by the ROBOTIS for MX-106 \cite{Robotis-MX106}.}
\label{fig:planned_velocities}
\end{figure}

The trajectories were not planned in accordance with the actuators limitations, which resulted in 
some of them reaching their velocity limit during the walk. This was especially true for the knee,
and ankle actuators which are the ones that experience the more torque during the walk. This is illustrated 
in figure \ref{fig:planned_velocities}.

\begin{figure}[h]
\centering
\includegraphics[width=6cm]{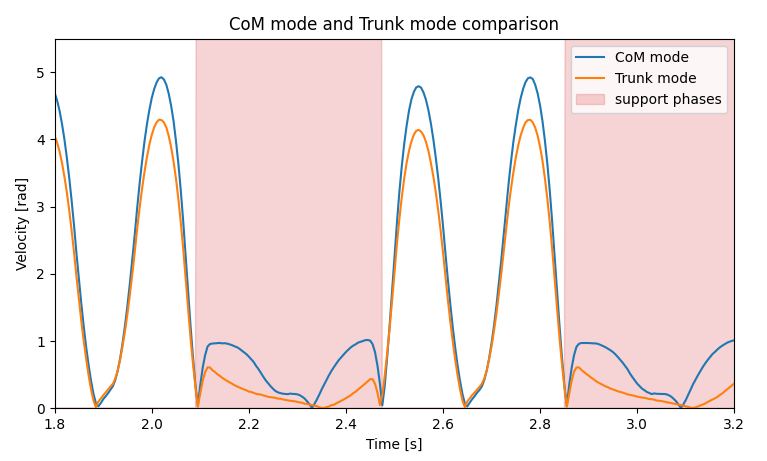}
\caption{Comparison of the planned velocity for the knee actuator during the CoM-walk (blue) and the walk in trunk mode (orange).}
\label{fig:knee_example}
\end{figure}

To avoid extra velocity solicitations of high torque-providing actuators, we decided to substitute the center 
of mass by the trunk position while resolving the whole-body kinematics. This "trunk mode" (as opposed
to the "CoM mode") is motivated by the fact that most of the mass is located in the trunk, and by the
fact that the constant CoM height constraint is the source the targeted velocity solicitations.
As a result, the target velocities  of the support leg actuators are lower (cf. figure \ref{fig:knee_example}),
which allows to stay in the operational range. 

Taking in account the true ability of the actuators when planning and controlling is really
important in a competitive scenario where the robot is pushed to its limits of operation.

The only feedback we currently use to ensure the stability of the robot is a detection of the support using the pressure sensors of the feet.
In the case of a light instability, the robot will take more time to touch the ground with the swing foot. Thus, we delay the trajectory reading
to ensure that the real state of the robot match the planned one. Another objective is thus to implement a proper model predictive control (MPC) 
to be resilient to greater instabilities.

Facing the limitations of the Sigmaban platform, in particular the difficulty to use the actuators with torque control, we are considering to move to a new architecture.

\section{Strategy and monitoring}
\label{sec:viewer_simulator}

Even if those problems are actually intrinsically mixed and should be ideally solved together, we decided to
split the game strategy into two main stages: the \textbf{kick strategy} which is used by a robot to decide what
kick should be used and where it should be targeted, and the \textbf{placement strategy} which is used to decide
which robot is going to perform the kick and where the other robots should be placed on the field.

\subsection{Kick strategy}

Four different kick motions are currently implemented in the robot: a forward kick
(\textit{classic}), a short forward kick (\textit{small}), a lateral kick (\textit{lateral})
and a diagonal one (\textit{diag}). For all of them, we estimated the optimal relative positioning,
and the distribution of kick length and angle. We also model the effect of kicking against the
grass blades as a reduction of 70\% of the kick length.
We first formalize the problem as a Markov Decision Process where the state $s$ is the position of the
ball on the field, the action $a$ the kick type and yaw, and the reward:

\begin{equation}
r
=
\begin{cases}
-t_k & \text{if the ball is in the goal}\\
-t_k -t_w(s, s') & \text{if the ball is on the field}\\
-t_p & \text{if the ball is out of the field}\\
\end{cases}
\end{equation}

Here, $t_k$ is the duration of a kick, $t_w$ is the time it takes to walk from $s$ to $s'$, and $t_p$ is
a penalty duration. 
We first use the \textit{Value Iteration} algorithm to (off-line) compute a
baseline value for each state, that can be interpreted as the (negative) time it takes to score a
goal from this state. 

On-line, we perform a (one step) tree-search to find the best action to perform, using an
augmented reward plus the baseline value of the next state. At this stage, the uncertainty of each possible kick
is still accounted for by sampling the kick length and angle from the estimated distribution.
The following extra information are added to the reward:

\begin{itemize}
    \item
    If the kick would \textbf{collide} an ally or an opponent, we recursively compute the reward for the two
    possible state using a collision probability,
    \item
    If we are \textbf{not allowed to score} (because of indirect free kick) and the kick would score, we add
    extra penalty,
    \item
    Instead of $t_w(s, s')$, we use the time it would take to the \textbf{closest ally} to walk to the ball,
    \item
    We add extra penalty if the ball would reach a position where the \textbf{opponents are closer to the ball},
    \item
    We add extra penalty if the robot is targeting a position that \textbf{obstructs our goals} to perform its kick.
\end{itemize}

\begin{figure}[h]
\centering
\includegraphics[width=12cm]{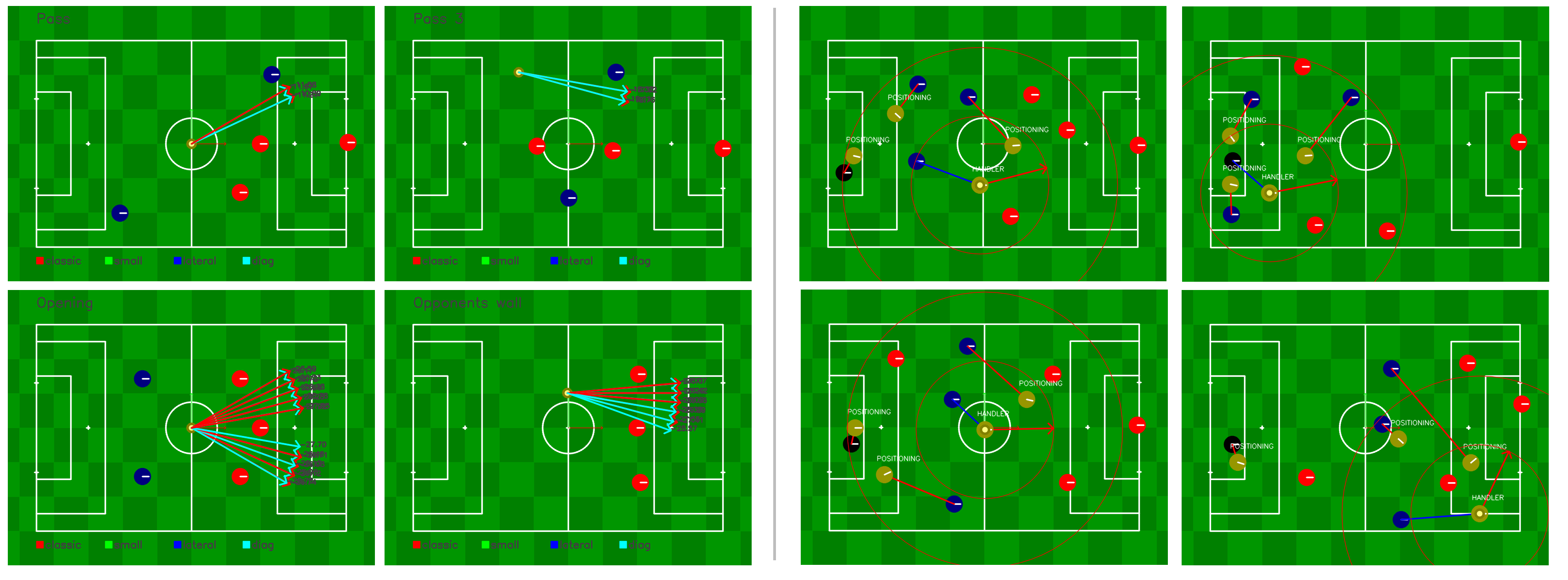}
\caption{
    On the left: examples of situations for the kicking strategy problem. On the right: situations
    for the placement strategy problem.
}
\label{fig:strategy}
\end{figure}

We used a graphical drag-and-drop representation as depicted on figure \ref{fig:strategy} to hand-tune the
reward according to our preferences and across a playbook of multiple standard situations.
The top 10\% of those possible actions are then passed to the footsteps
generator, which uses its value function (see section \ref{sec:footsteps}) in inference to estimate the number of
steps required for each action and select the best one accordingly.

\subsection{Placement strategy}

The robots are distributed autonomous agents that need to agree on a common team strategy.
To simplify this problem, we leverage the WiFi communication between robots to elect a Captain that
is responsible for computing and broadcasting a global team roles assignment.
The captain listens to robots information such as estimated location on field, location of the ball and allies opponents detection (Protobuf-encoded packets over UDP broadcast). Next, the captain applies clustering algorithm to aggreagate robots observations and infer common opponents detection. A kicker is selected among robots using a simple scoring (bonus for previous kicker to avoid oscillation, distance to ball and robot orientation). The remaining robots are placed over a set of interesting positions (empirically selected).

\subsection{Monitoring and replay}

\begin{figure}[h]
\centering
\includegraphics[width=4cm]{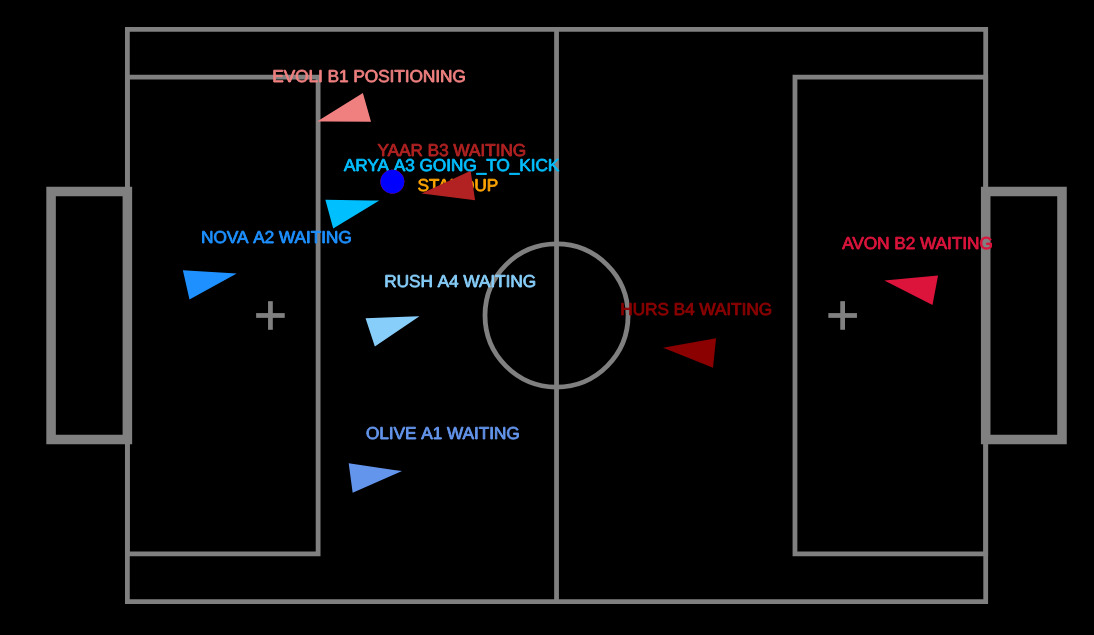}
\caption{Monitoring system}
\label{fig:ksv}
\end{figure}

We designed a new monitoring system written entirely in Python for convenience. It listens over the network for
the \textit{Game Controller}, robot communication, records a video from an external webcam to keep a ground truth
overview. Such a tool allows real time debugging during development
phases, but also to replay a game and analyze the strategy (figure \ref{fig:ksv}).

%
% ---- Bibliography ----
%
% BibTeX users should specify bibliography style 'splncs04'.
% References will then be sorted and formatted in the correct style.
%
\bibliographystyle{splncs04}
\bibliography{ChampionPaper2023}
\end{document}